%% file: main.tex
\documentclass[10pt,twocolumn,letterpaper]{article}

\usepackage[pagenumbers]{cvpr} 

\usepackage{graphicx}
\usepackage{amsmath}
\usepackage{amssymb}
\usepackage{booktabs}
\usepackage{kotex}
\usepackage{xcolor}
\usepackage{xspace}
\usepackage{multirow}
\usepackage[T1]{fontenc}
\usepackage{pifont}
\usepackage{comment}
\usepackage{wrapfig}
\usepackage{caption}
\usepackage{subcaption}
\usepackage{mathtools}
\usepackage[accsupp]{axessibility}  

\input{math_commands.tex}

\usepackage[pagebackref,breaklinks,colorlinks]{hyperref}

\usepackage[capitalize]{cleveref}
\crefname{section}{Sec.}{Secs.}
\Crefname{section}{Section}{Sections}
\Crefname{table}{Table}{Tables}
\crefname{table}{Tab.}{Tabs.}

\newcommand{\myparagraph}[1]{\vspace{0pt}\noindent{\bf #1}}
\newcommand{\TODO}[1]{\textcolor{red}{TODO: #1}}

\newcommand\CAMFull{$\textit{CAM}_\textit{full}$\xspace}
\newcommand\CAMPartial{$\textit{CAM}_\textit{partial}$\xspace}
\newcommand\boostrelu{$\text{BoostLU}$\xspace}

\def\cvprPaperID{10120} 
\def\confName{CVPR}
\def\confYear{2023}

\begin{document}

\title{Bridging the Gap between Model Explanations in \\Partially Annotated Multi-label Classification}

\author{Youngwook Kim$^{1}$
~~~~~~~~~
Jae Myung Kim$^{2}$
~~~~~~~~~
Jieun Jeong$^{1,3}$
\\
Cordelia Schmid$^{4}$
~~~~~~~~~
Zeynep Akata$^{2,5}$ 
~~~~~~~~~
Jungwoo Lee$^{1,3}\thanks{Corresponding author.}$
\\
~\\
\small{
$^1$Seoul National University\quad
$^2$University of T\"{u}bingen\quad
$^3$HodooAI Lab\quad
} \\
\small{
$^4$Inria, Ecole normale sup\'erieure, CNRS, PSL Research University\quad
$^5$MPI for Intelligent Systems\quad
}
}
\maketitle

\begin{abstract}
   Due to the expensive costs of collecting labels in multi-label classification datasets, partially annotated multi-label classification has become an emerging field in computer vision. One baseline approach to this task is to assume unobserved labels as negative labels, but this assumption induces label noise as a form of false negative. To understand the negative impact caused by false negative labels, we study how these labels affect the model's explanation. We observe that the explanation of two models, trained with full and partial labels each, highlights similar regions but with different scaling, where the latter tends to have lower attribution scores. Based on these findings, we propose to boost the attribution scores of the model trained with partial labels to make its explanation resemble that of the model trained with full labels. Even with the conceptually simple approach, the multi-label classification performance improves by a large margin in three different datasets on a single positive label setting and one on a large-scale partial label setting.
   Code is available at \url{https://github.com/youngwk/BridgeGapExplanationPAMC}.
\end{abstract}


\input{01intro}

\input{02related}

\input{03camanalysis}

\input{04method}

\input{05experiments}

\input{06conclusion}

\clearpage

{\small
\bibliographystyle{ieee_fullname}
\bibliography{main}
}

\end{document}


\title{Supplementary Material for "Bridging the Gap between Model Explanations in \\Partially Annotated Multi-label Classification"}

\author{Youngwook Kim$^{1}$
~~~~~~~~~
Jae Myung Kim$^{2}$
~~~~~~~~~
Jieun Jeong$^{1,3}$
\\
Cordelia Schmid$^{4}$
~~~~~~~~~
Zeynep Akata$^{2,5}$ 
~~~~~~~~~
Jungwoo Lee$^{1,3}\thanks{Corresponding author.}$
\\
~\\
\small{
$^1$Seoul National University\quad
$^2$University of T\"{u}bingen\quad
$^3$HodooAI Lab\quad
} \\
\small{
$^4$Inria, Ecole normale sup\'erieure, CNRS, PSL Research University\quad
$^5$MPI for Intelligent Systems\quad
}
}

\maketitle

\begin{alphasection}

\section{Hyperparameter sensitivity}
In this section, we check the hyperparameter sensitivity of our proposed BoostLU. All experiments are conducted for LL-Ct + BoostLU in a COCO dataset. Figure \ref{fig:alpha} shows the experimental results for various $\alpha$ with fixed $\beta=0$. Note that $\alpha=1$ refers to the case of the original LL-Ct since positive attribution scores are not scaled. When $\alpha$ exceeds 1, 
performance rises as the attribution score damaged by the false negative begins to be compensated.
The performance gradually increases and peaks at $\alpha=5$.
Figure \ref{fig:beta} shows the performance trend for various $\beta$ with fixed $\alpha=5$. 
According to the results, the value of $\beta$ does not significantly affect the model's performance.
These two figures represent that our BoostLU is generally robust to its hyperparameters $\alpha$ and $\beta$.
\begin{figure}[t]
    \centering
    \includegraphics[width=0.9\linewidth]{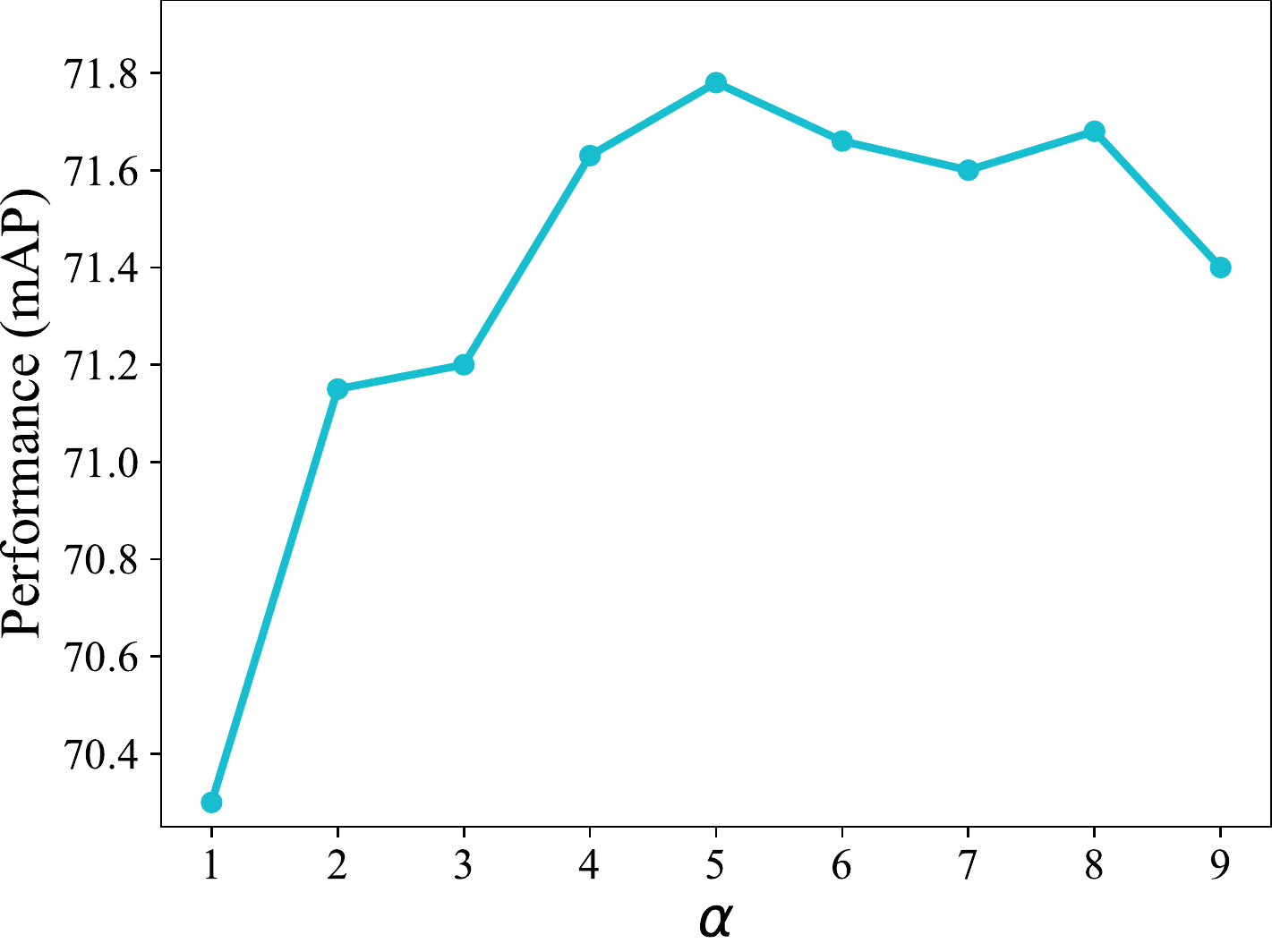}
    \caption{\textbf{Hyperparameter sensitivity with respect to $\alpha$.} 
    } 
    \label{fig:alpha}
\end{figure}

\begin{figure}[t]
    \centering
    \includegraphics[width=0.9\linewidth]{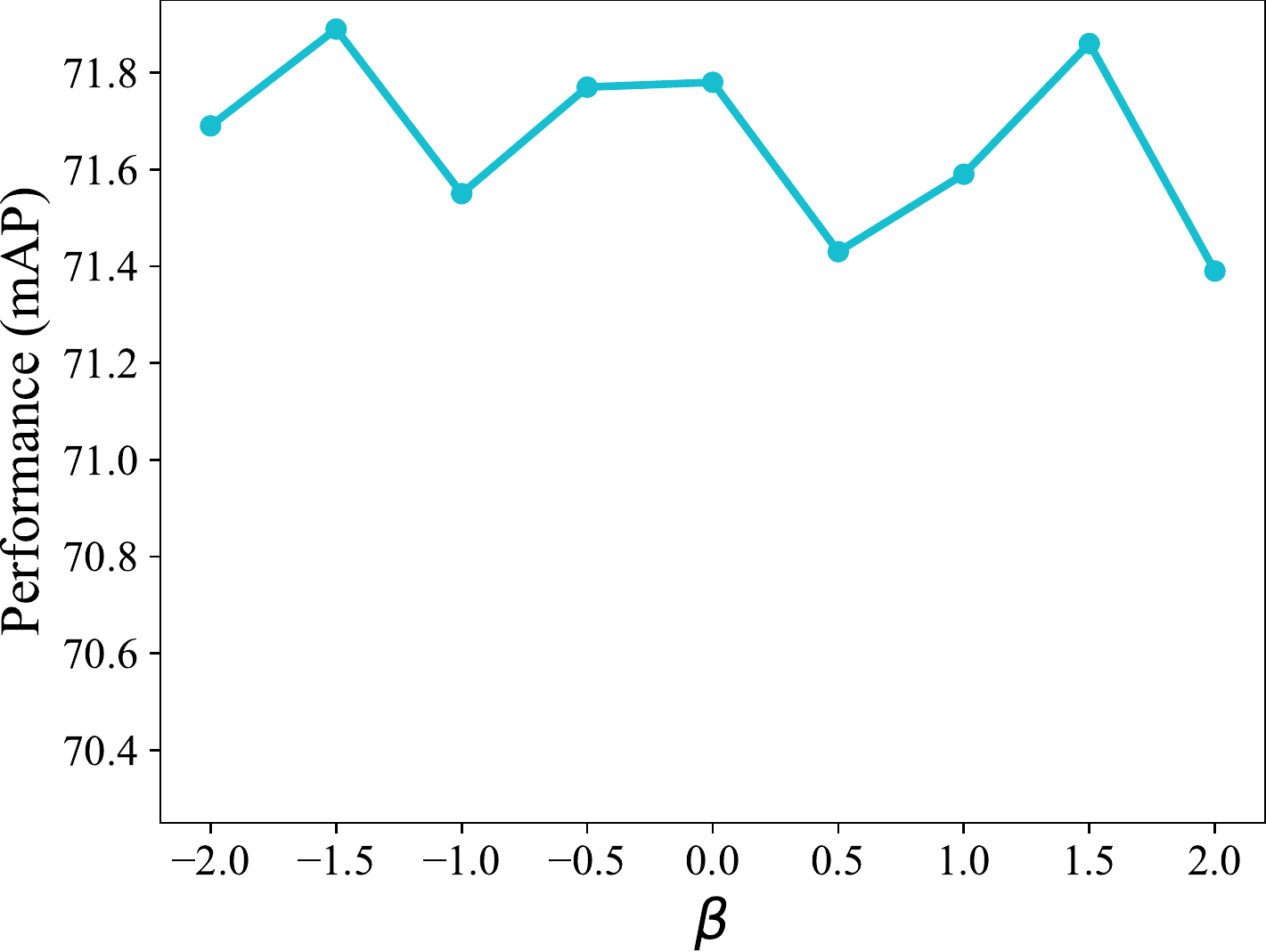}
    \caption{\textbf{Hyperparameter sensitivity with respect to $\beta$.} 
    } 
    \label{fig:beta}
\end{figure}

\section{Additional discussion about Table 1}
When LL-R and BoostLU are used in training, but BoostLU is not used in inference (fifth row), the network is \textit{optimized} via BoostLU-activated attribution scores. So after training, pre-activated attribution scores for positive labels would become smaller, even though false negatives are further alleviated.
It leads to worse model performance, even lower than when BoostLU is not used in training, but only LL-R is used in training (fourth row). From this result,
we can confirm the importance of applying BoostLU in inference to obtain performance gain.

\end{alphasection}

%% file: math_commands.tex
\usepackage{amsmath,amsfonts,bm}

\def\eqref#1{equation~\ref{#1}}

\def\1{\bm{1}}

\def\vx{{\bm{x}}}
\def\vy{{\bm{y}}}

\def\mF{{\bm{F}}}

\def\mM{{\bm{M}}}

\def\mW{{\bm{W}}}

\DeclareMathAlphabet{\mathsfit}{\encodingdefault}{\sfdefault}{m}{sl}
\SetMathAlphabet{\mathsfit}{bold}{\encodingdefault}{\sfdefault}{bx}{n}

\def\gD{{\mathcal{D}}}

\def\gI{{\mathcal{I}}}

\def\gY{{\mathcal{Y}}}



%% file: 01intro.tex
\section{Introduction}
\label{sec:intro}
Multi-label image classification is the task of predicting all labels corresponding to a given image.
Since web-crawled images often contain multiple objects/concepts \cite{imagenet1, imagenet2, beyer2020we, rajeswar2022multi}, the importance of this task is rising.
However, it faces a significant issue of huge annotation costs.
We need C binary labels for each training image to provide exhaustive annotation for a model that classifies images into C categories.
It acts as a severe obstacle to scaling multi-label classification datasets.

For this reason, partially annotated multi-label classification \cite{partial_label_2019, imcl, kundu2020exploiting, single_positive_label, kim2022large, p-asl} has recently become an actively studied topic.
In this setting, instead of exhaustive annotation, only a few categories are labeled for each training image.
We can effectively reduce the burden of annotation by adopting partial annotation strategies.

\begin{figure}[t]
    \centering
    \includegraphics[width=\linewidth]{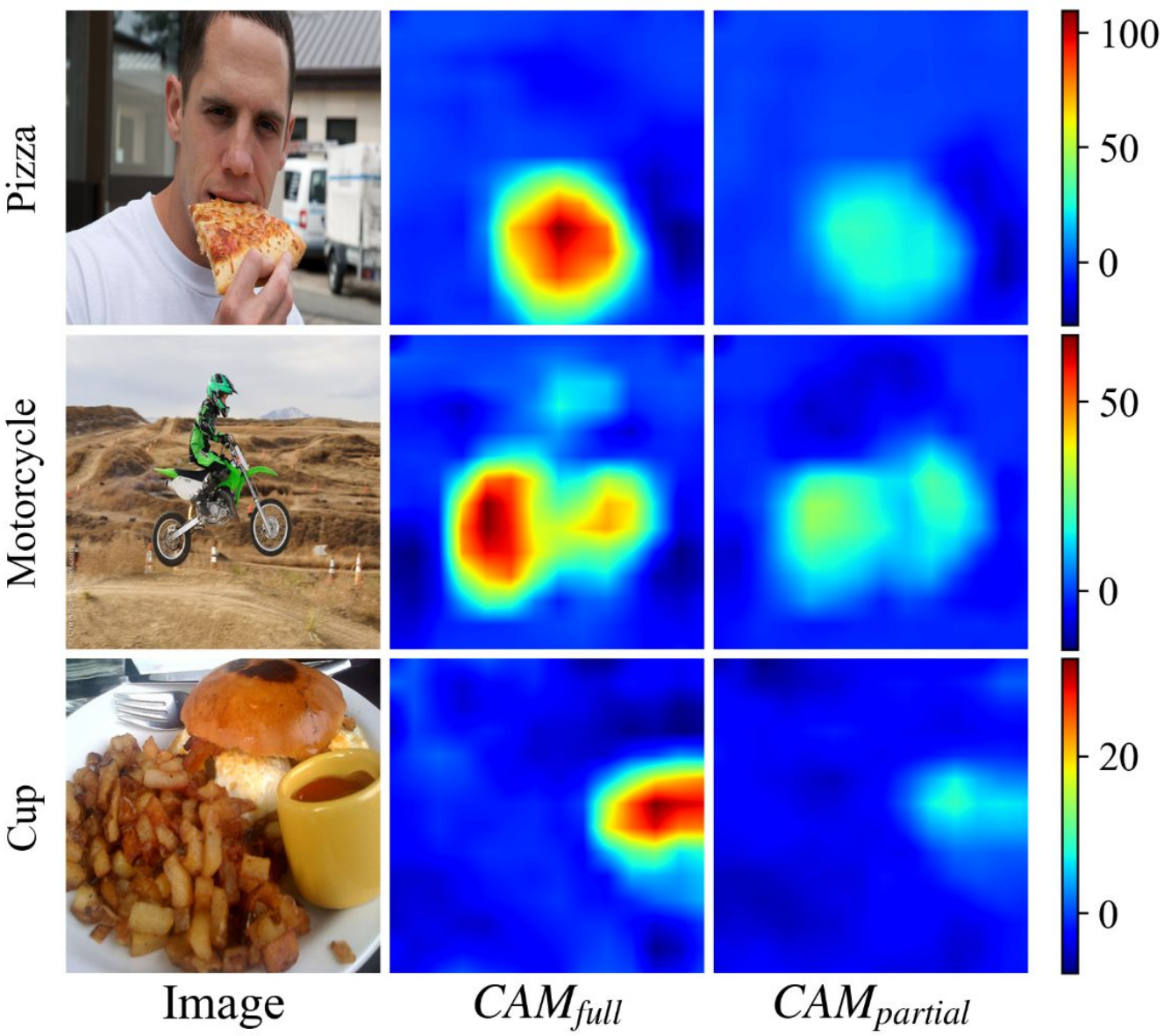}
    \caption{\textbf{CAM Observation.} We compare the class activation map (CAM) output from two multi-label classification models: one trained with full labels (\CAMFull) and the other trained with partial labels and AN assumption (\CAMPartial). We observe that the overall structure of \CAMPartial is not much affected by the noisy false negative labels during training. This observation motivates us to make \CAMPartial similar to \CAMFull by boosting its relatively large attribution scores. Best viewed in color.
    } 
    \vspace{-10pt}
    \label{fig:camvis}
\end{figure}

One baseline approach for solving a partially annotated multi-label classification task is assuming unobserved labels as negative labels (Assume Negative, AN) \cite{wsml1, wsml2, chen2013fast, wang2014binary}. It is a reasonable assumption since most labels are negative labels in the multi-label scenario \cite{asl}. 
However, this assumption causes label noise in a form of false negatives since the actually positive but unannotated labels are incorrectly assumed to be negative.
Since this label noise perturbs the learning process of the model \cite{zhang2021understanding, chen2021beyond, memorization, mentornet}, recent studies on a partially annotated multi-label classification focus on suppressing the influence of label noise 
by ignoring or correcting the loss of samples that are likely to be false negatives\cite{kim2022large, p-asl}.


Aside from recent research directions, we delve into “how” false negative labels influence a multi-label classification model. We conduct control experiments with two models. One is the model trained with partial labels and AN assumption where false negative labels exist. The other is the model trained with full annotations and thus trained without false negatives.
We compare the class activation map (CAM) \cite{cam} output between the two models to see the difference in how each model understands the input image and makes a prediction result.


Figure \ref{fig:camvis} shows that a model trained with false negatives still highlights similar regions to one trained with full annotation. However, the attribution scores in the highlighted areas are much small. 
This observation leads us to think that if we scale up the damaged score of the highlighted region in the model trained with false negatives, the explanation of this model will become similar to that of the model trained with full annotation.





To this end, we introduce a simple piece-wise linear function, named \boostrelu, that bridges the gap between the explanation of two models trained with false negatives and with full annotation each. Concretely, we use the modified CNN model to get CAM during the forward pass directly\cite{acol}, and the logit in the modified CNN model is the mean of attribution scores of CAM. The \boostrelu function is applied element-wisely to the CAM output of the modified CNN to boost the scores of the highlighted regions, thereby compensating for the decrease of attribution scores in CAM caused by false negatives. It increases the logit value for positive labels and thus makes a better prediction.
Furthermore, when we combine \boostrelu with the recently proposed methods \cite{kim2022large} that explicitly detect and modify false negatives during training, it helps to detect false negatives better, thus leading to better performance. 
As a result, we achieve state-of-the-art performance on PASCAL VOC \cite{pascalvoc}, MS COCO \cite{coco}, NUSWIDE \cite{nuswide}, and Openimages V3 \cite{openimages} datasets in a partial label setting.

We summarize the contributions of this paper as follows.
\begin{enumerate}
    \item We analyze how the false negative labels affect the explanation of the model in a partially annotated multi-label classification scenario.
    \item We propose a simple but effective function, named \boostrelu, that compensates for the damage of false negatives in a multi-label classification model with little extra computational cost.
    \item When applied during inference, \boostrelu boosts the baseline method (AN)'s test performance without additional training.
    \item Combined with recent methods of detecting and modifying false negatives during training, \boostrelu boosts the state-of-the-art performance on single positive and large-scale partial label settings.
\end{enumerate}

%% file: 02related.tex
\section{Related Works}
\myparagraph{Partially annotated multi-label classification.}
One primary stream to solve the partially annotated multi-label classification problem is to view unobserved labels as \textit{missing labels}.
Earlier works tackled this problem by solving matrix completion \cite{cabral2011matrix, matrixcompletion2, matrixcompletion3} or employing the Bayesian model \cite{prob1, prob2}. However, these works require loading all data into memory at once, thus making it infeasible to train deep neural networks. 
Curriculum labeling \cite{partial_label_2019} proposed a bootstrapping strategy using model prediction. IMCL \cite{imcl}, SE \cite{kundu2020exploiting}, and SST \cite{sst} exploited label correlation and image similarity to generate regularization losses or pseudo-labels for missing labels. SARB \cite{sarb} performed a category-wise mixup on feature space between labeled and unlabeled images to propagate information into missing labels. Zhou et al. \cite{zhou2022acknowledging} proposed entropy maximization loss that suppresses gradients from missing labels to promote learning from observed labels. 

Since a significant part of labels is negative in a multi-label setting \cite{asl}, there is another stream to treat unobserved labels as negatives and try to lessen the harmful impact of false negatives. In other words, it views unobserved labels as \textit{noisy labels}.
ROLE \cite{single_positive_label} proposed to estimate unobserved labels while simultaneously regularizing the estimation with an average number of positive labels online. 
Kim et al. \cite{kim2022large} observed the memorization effect \cite{memorization} in a noisy multi-label classification setting that the model learns from clean labels first. Thus false negative labels are likely to show large loss values during training.
Then they suggested three methods, LL-R, LL-Ct, and LL-Cp, that prevent false negatives from being memorized by rejecting, temporally correcting, and permanently correcting samples with large losses, respectively.
P-ASL \cite{p-asl} assigned different scaling rates between annotated negatives and assumed negatives. It also ignored losses from categories with high prediction scores or label prior values.
In this work, we look at false negatives differently and study their effect on model explanation.
 

\myparagraph{Class activation mapping.} Class activation mapping (CAM) \cite{cam} provides information about where the classification model is attending to generate prediction scores. There are several follow-up works, including Grad-CAM \cite{gradcam}, which generates model-agnostic attention maps, and CALM \cite{calm}, which strengthens the interpretability of attention maps.

Since CAM provides localization ability to classification models,
it has been widely used for various vision tasks, such as weakly supervised object localization \cite{xie2022c2am, wu2022background, cole2022label, choe2020evaluating} and weakly supervised semantic segmentation \cite{xie2022c2am, liu2022adaptive, lee2021railroad, zhang2020causal, lee2021anti}. Recently, Zhang et al. \cite{zhang2022learnfromall} utilized CAM in facial expression recognition in the presence of noisy labels. They found that the model trained with noisy labels highlights only part of the features and suggested a random masking strategy to prevent memorizing partial features.
Although there is a similarity in that they inspected the CAM output of the model in noisy label situations, our work is different since we focus on the noisy multi-label classification setting with another type of noise.

%% file: 03camanalysis.tex
\section{Preliminary}

\begin{figure*}[t]
    \centering
    \begin{subfigure}[b]{0.325\textwidth}
        \centering
        \includegraphics[width=\textwidth]{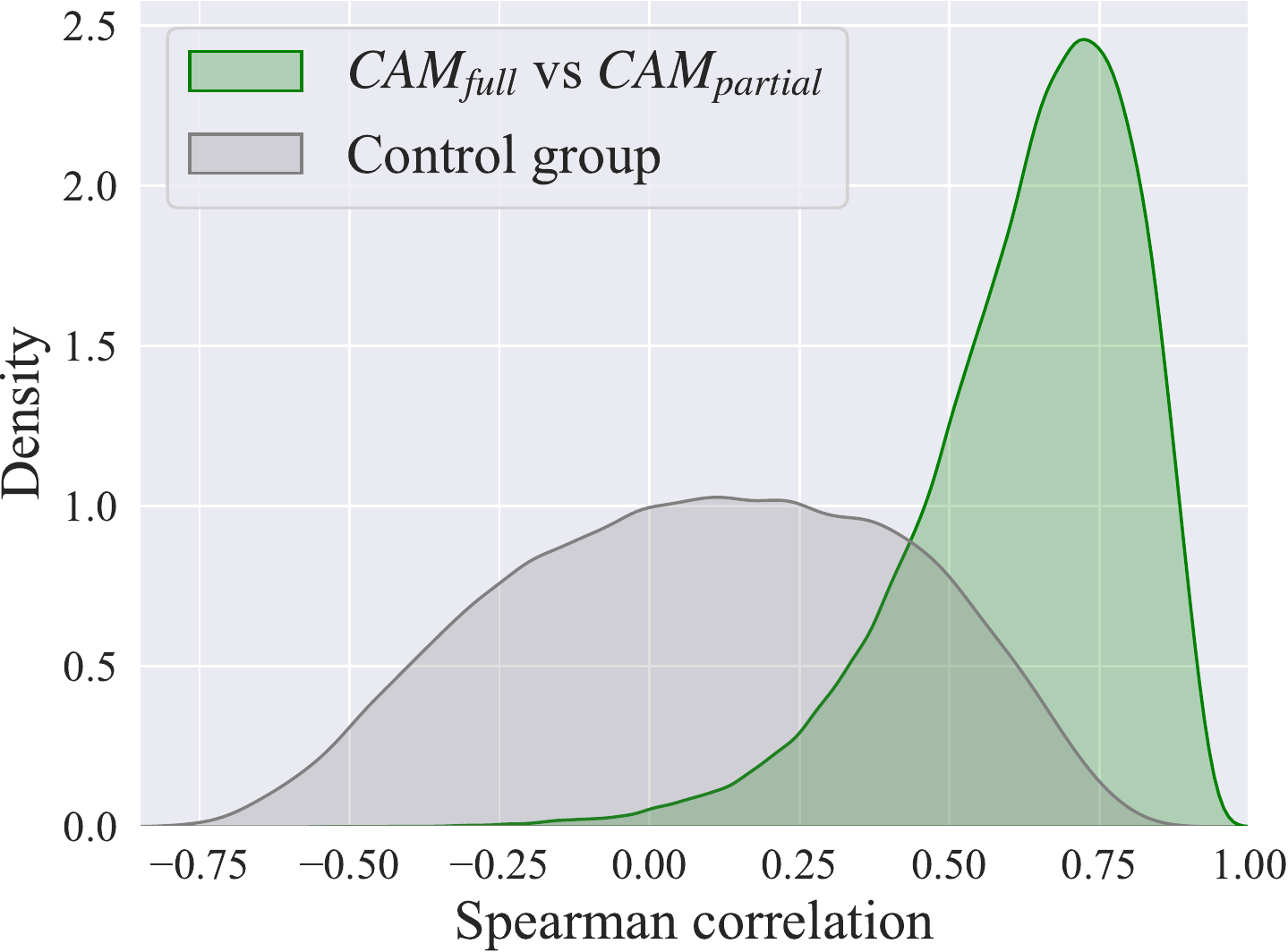}
        \caption{Similarity of CAM by Spearman correlation}
        \label{fig:spearman}
    \end{subfigure}
    \hfill
    \begin{subfigure}[b]{0.325\textwidth}
        \centering
        \includegraphics[width=\textwidth]{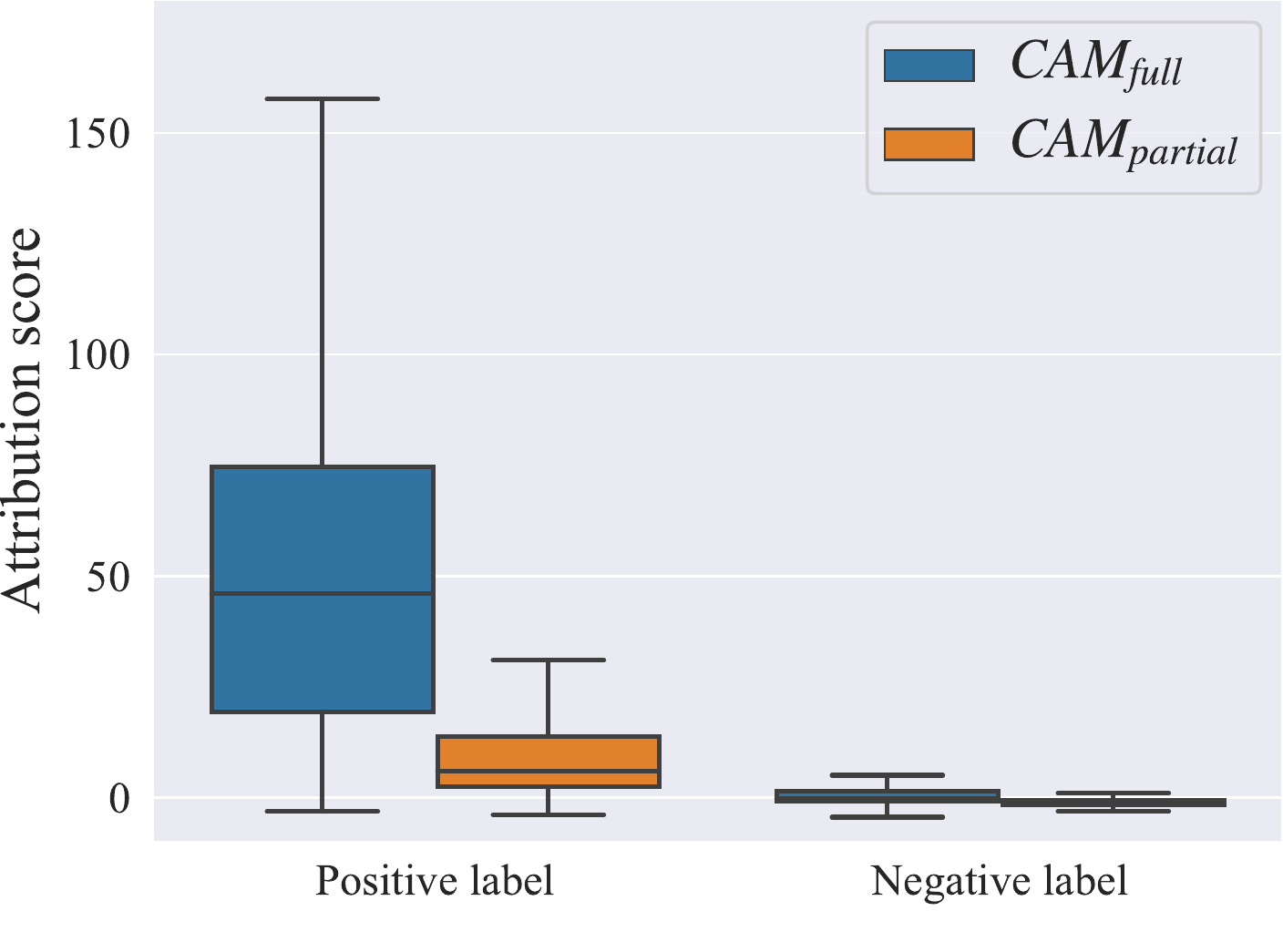}
        \caption{Top-ranking attribution score}
        \label{fig:cammax}
    \end{subfigure}
    \hfill
    \begin{subfigure}[b]{0.325\textwidth}
        \centering
        \includegraphics[width=\textwidth]{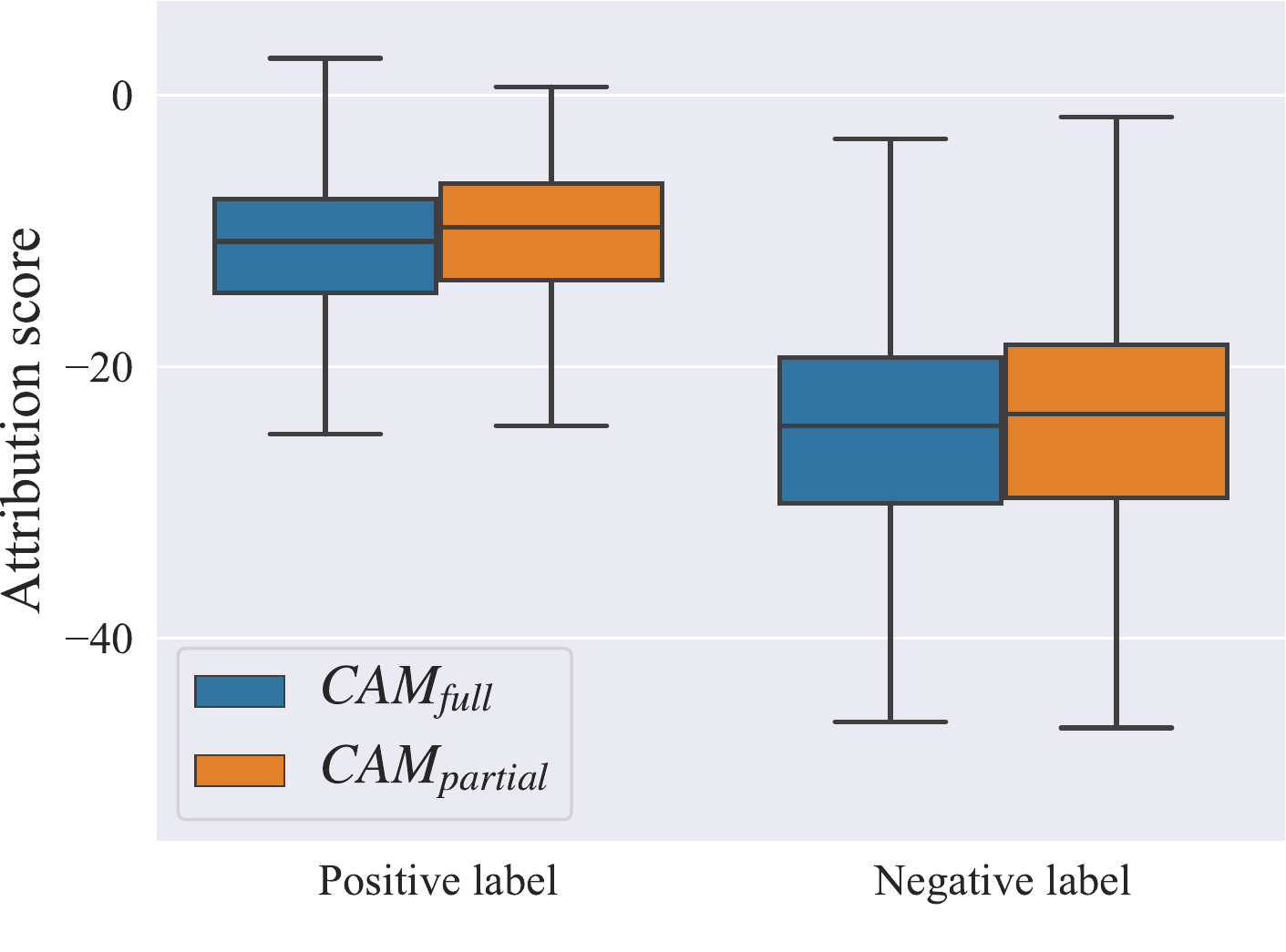}
        \caption{Bottom-ranking attribution score}
        \label{fig:cammin}
    \end{subfigure}
    \vspace{-5pt}
    \caption{\textbf{CAM Analysis on COCO test set.} (a): Distribution of Spearman correlation coefficients between \CAMFull and \CAMPartial from the same image. Overall positive correlation implies that \CAMPartial has a structure similar to \CAMFull. 
    \slash\ (b), (c): Boxplot of the average of top\slash bottom 5\% of attribution scores, respectively.
    The damage of false negative labels to the model mainly lowers the upper attribution scores for positive labels
    while maintaining its overall structure in CAM. 
    } 
    \vspace{-5pt}
    \label{fig:camanalysis}
\end{figure*}

This section introduces the formal definition of a partially annotated multi-label classification in \S\ref{sec:problemdef}. Next, we briefly summarize the class activation map (CAM) in \S\ref{sec:recapcam}.
\subsection{Problem Definition}
\label{sec:problemdef}
We aim to train a multi-label classification model with dataset $\gD$ consisting of pairs of input image $\vx$ and partially annotated label $\vy$. Each category can have three kinds of labels: 0, 1, and $\phi$. In other words, $\vy \in \gY = \{0,1,\phi\}^C$ where $\phi$ indicates the absence of annotation and $C$ is the number of total categories.
Denote the index set of positive labels, negative labels, and unannotated labels as $\gI^{\,p}$, $\gI^{\,n}$, and $\gI^{\,\phi}$, respectively.
We study the setting where labels are sparsely annotated, i.e., $|\gI^{\,p}| + |\gI^{\,n}| \ll |\gI^{\,\phi}|$. 

A straightforward approach to train the model given partial labels is to treat unannotated labels by assuming negative (AN) and use binary cross-entropy as a loss function:
%

\begin{equation}
    \mathcal{L}_{AN}=\frac{1}{C}\left[\,\,
    \sum_{i\, \in\, \gI^{\,p}}  \, \mathcal{L}_{+} 
    +  \sum_{i\, \in\,
    \gI^{\,n} \cup\, \gI^{\,\phi}} \, \mathcal{L}_{-}\,
    \right] \label{eq:anloss}
\end{equation}
%
where $\mathcal{L}_{+}=-\log(\sigma(g_i))$, 
$\mathcal{L}_{-}=-\log(1 \! - \! \sigma(g_i))$
and $g_i$ is a logit for $i$-th category. However, labels whose true label is positive but unannotated are incorrectly assumed to be negative and become false negatives. Denote the index set of true negative and false negative labels as $\gI^{\,tn}$ and $\gI^{\,fn}$, then $\gI^{\,n} \,\cup\, \gI^{\,\phi} = \gI^{\,tn} \,\cup\, \gI^{\,fn}$. We set the approach of training the model with Equation (\ref{eq:anloss}) as the baseline method and investigate the influence of false negatives on the multi-label classification model.

\subsection{Recap CAM}
\label{sec:recapcam}
Most CNN architectures consist of several convolution layers (Convs), followed by a Global Average Pooling (GAP) layer \cite{lin2013network} and a fully connected (FC) layer. 
Let the last convolutional feature map be $\mF \in \mathbb{R}^{H \times W \times D}$, and a weight matrix of the FC layer be $\mW \in \mathbb{R}^{C \times D}$ where $(H,W)$ and $D$ are the spatial size and channel size of the feature map, respectively. 
We can obtain the class activation map (CAM) \cite{cam} for class c ($\mM_c$) by

\begin{equation}
    \mM_c = \sum_{d=1}^D \mW_{cd} \mF_d \,\, ,
    \label{eq:CAM}
\end{equation}
where $\mF_d$ denotes $d$-th channel of $\mF$. $\mM_c$ explains the model's prediction by attributing scores on each pixel. 

Instead of performing post-processing to get CAM as in Equation (\ref{eq:CAM}), we can directly get CAM during the forward pass by reordering the last two layers from Convs-GAP-FC to Convs-1x1Conv-GAP where 1x1Conv is the one-by-one convolutional layer with the weight $\mW$ \cite{acol}. The output feature maps of 1x1Conv become the same as $\mM$, and the logit $g_c$ becomes
\begin{equation}
    g_c = \frac{1}{HW}\sum_{i=1}^H\sum_{j=1}^W (\mM_c)_{ij}  \,\, . \label{eq:camtologit}
\end{equation}
Thus, we can interpret each element $(\mM_c)_{ij}$ as an \textit{attribution score} at spatial location $(i,j)$ contributing to the logit for class $c$. For the following sections, we utilize this modified architecture to facilitate the application of our method.



\section{Impact of False Negatives on CAM}
\label{sec:camanalysis}
It is well known that neural networks can memorize wrong labels due to their large model capacities \cite{zhang2021understanding}.
Likewise, if we train a multi-label classification model with AN loss (Equation (\ref{eq:anloss})) when given partial labels, the model is damaged by memorizing false negative labels \cite{kim2022large}. It results in poor performance compared to the model trained with full labels, which false negatives have not influenced.

To better understand why the model trained with partial labels performs less than that with full labels, we analyze the behavioral difference between these two models. Concretely, we use a class activation map (CAM) \cite{cam} to explain each model's prediction and compare the explanation results. 
We train two multi-label classification models on a COCO dataset \cite{coco} with the same CNN architecture ResNet-50 \cite{resnet}: 
one model with full labels using binary cross entropy loss and the other with partial labels using AN loss (Equation (\ref{eq:anloss})).
We denote the CAM output from each model as \CAMFull and \CAMPartial, respectively. 

To analyze the explanation of these two models, we first compute the Spearman correlation between \CAMFull and \CAMPartial on positive labels.
We show the distribution of the correlation values on the test set in Figure \ref{fig:spearman}. For comparison, we consider a 2D Gaussian image centered at the midpoint and calculate the Spearman correlation coefficient between this Gaussian image and \CAMFull. We observe that there is mainly a positive correlation between \CAMFull and \CAMPartial, while the correlation of the control group is distributed widely but mostly around zero. 
It implies that the overall structure (i.e., the attribution ranking among pixels) of \CAMPartial is preserved despite the influence of false negative labels, therefore having a high Spearman correlation with \CAMFull. We can also visually inspect the similar structure between \CAMPartial and \CAMFull in Figure \ref{fig:camvis}, where both CAMs highlight similar regions. 

Since we know that the overall structure is similar between \CAMFull and \CAMPartial, we then compare the range of attribution scores between \CAMFull and \CAMPartial. Concretely, we compute the mean of the highest 5\% of attribution scores and the lowest 5\%, respectively, for each CAM and summarize the distribution of these values on the test set in Figure \ref{fig:cammax} and \ref{fig:cammin}. Note that we take an average of 5\% of scores to reduce the effect of outliers.
We observe that top-ranking attribution scores of \CAMPartial from positive labels drop sharply compared to \CAMFull, while these scores from negative labels remain similar. Also, there is little difference in bottom-ranking attribution scores between \CAMFull and \CAMPartial, both on positive and negative labels.
It implies that false negatives mainly affect the model's understanding in regions with relatively high attribution scores, especially for positive labels. Consequently, the decrease of attribution scores at specific regions in CAM leads to a decrease in the logit value (since logit is the average of attribution scores on CAM as in Equation (\ref{eq:camtologit})), making the model predicts a lower score for the positive category. 
The change of gradient during training can explain the occurrence of this phenomenon.

\myparagraph{Gradient analysis.}
In Equation (\ref{eq:anloss}), recall that the BCE loss is $\mathcal{L}_{+}$ with a positive target and $\mathcal{L}_{-}$ with a negative one. Their gradients with respect to the logit $g$ are 
\begin{equation}
    \frac{\partial \mathcal{L}_{+}}{\partial g} = 
    \sigma(g)-1, \,\,\frac{\partial \mathcal{L}_{-}}{\partial g} = 
    \sigma(g) \,\, .
    \label{eq:dldg}
\end{equation}
For a training image $\vx$, 
the gradient difference on the logit $g$ between partial label (with AN assumption) and full label case is given by
%
\begin{equation}
\begin{aligned}
    &\frac{1}{C}\left[\,
    \sum_{i \,\in\, \gI^{\,p}} \frac{\partial \mathcal{L}_{+}}{\partial g_i} + \sum_{i \,\in\, \gI^{\,fn}} \frac{\partial \mathcal{L}_{-}}{\partial g_i}
    + \sum_{i \,\in\, \gI^{\,tn}} \frac{\partial \mathcal{L}_{-}}{\partial g_i}
    \right] \\
    &- \frac{1}{C}\left[\,
    \sum_{i \,\in\, \gI^{\,p}} \frac{\partial \mathcal{L}_{+}}{\partial g_i} + \sum_{i \,\in\, \gI^{\,fn}} \frac{\partial \mathcal{L}_{+}}{\partial g_i}
    + \sum_{i \,\in\, \gI^{\,tn}} \frac{\partial \mathcal{L}_{-}}{\partial g_i}
    \right]\\
    &=\frac{1}{C}\left[ \,
    \sum_{i \,\in\, \gI^{\,fn}} (\frac{\partial \mathcal{L}_{-}}{\partial g_i} - \frac{\partial \mathcal{L}_{+}}{\partial g_i})
    \right]
    \,=\,\,\frac{|\gI^{\,fn}|}{C} \,\, .
    \label{eq:gradientanalysis}
\end{aligned}
\end{equation}
Equation (\ref{eq:gradientanalysis}) shows that the logit receives more gradients proportional to the number of false negative labels on a partial label setting. 
Therefore, 
as training progresses, the additional gradients from false negatives are gradually accumulated in the logit, making the logit smaller than the model trained on full labels. Since the logit is equal to the average of CAM, the attribution scores of \CAMPartial become lower than that of  \CAMFull. 



%% file: 04method.tex
\section{Proposed Method}

In this section, we propose a conceptually simple but effective method to make the model trained with partial labels resemble the model trained with full labels by mimicking the explanation. We propose a function \boostrelu devised to compensate for the damaged attribution score of the explanation due to false negatives in \S\ref{sec:boostrelu}. We then introduce three scenarios that utilize our function through \S\ref{sec:usage1} $\sim$ \S\ref{sec:usage3}.

\subsection{BoostLU}
\label{sec:boostrelu}

From the modified CNN architecture described in \S\ref{sec:recapcam}, define convolutional layers (Convs-1x1Conv) as $\Phi$.
Given an input image $\vx$, its class activation map (CAM) is $\mM = \Phi (\vx)$.
Our goal is to make the explanation $\mM$ of the model trained with partial labels closer to the explanation of the model trained with full labels, even if we do not have access to the full labels, thus improving the prediction performance. 

\begin{figure}[t]
    \centering
    \includegraphics[width=\linewidth]{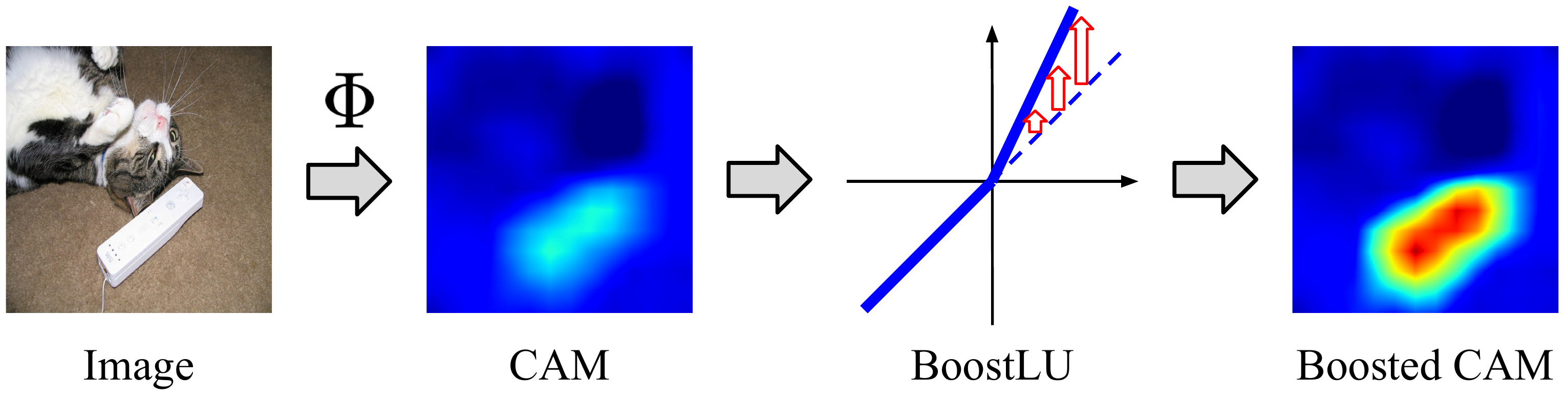}
    \vspace{-15pt}
    \caption{\textbf{Schematic diagram of applying \boostrelu.} \boostrelu is applied to the model's CAM output element-wisely to compensate for the attribution scores damaged by false negative labels.} 
    \vspace{-10pt}
    \label{fig:boostrelu}
\end{figure}

In the previous section, we observe that when a multi-label classification model is trained with AN loss, the way the model understands images is damaged by false negatives. However, we also find that this damage is mainly focused on a drop in the relatively high attribution scores while the overall spatial structure of CAM is preserved.
Based on these findings, we conjecture that if the damaged high attribution scores are scaled up in the model trained with partial labels, \CAMPartial will become similar to \CAMFull. 
To achieve this, we devise a piece-wise linear function that boosts the attribution scores that are above a certain threshold:

\begin{equation}
    f(x) =
    \begin{cases}
        \alpha x + (1 - \alpha)\beta, & x \geq \beta \\
        x, & x < \beta \,\, ,
    \end{cases}
    \label{eq:boostrelu_original}
\end{equation}
where $\alpha$ is a scaling factor with $\alpha > 1$, and $\beta$ is a threshold determining whether to boost the score. 
Since top-ranking attribution scores on CAM tend to have large positive values for positive labels and around zero for negative labels
(as seen in Figure \ref{fig:cammax}), we search for the values of $\beta$ around zero. Since we empirically observe no significant difference in model performance for different $\beta$ (these results are reported in the Appendix), we only consider the simplest case of $\beta = 0$. 
Then we can rewrite Equation (\ref{eq:boostrelu_original}) in a ReLU-like form as
%
\begin{equation}
    \text{\boostrelu}(x) = max(x, \alpha x) \,\, .
    \label{eq:boostrelu_beta0}
\end{equation}
%
By applying \boostrelu to each element of CAM, as illustrated in Figure \ref{fig:boostrelu},
\boostrelu boosts positive attribution scores by $\alpha$ times, which are the main target to be damaged by false negatives, while maintaining the negative scores unchanged.
These selectively boosted attribution scores are aggregated through the GAP layer to produce a logit value as 


\begin{equation}
    g(\vx) = (\text{GAP} \circ \text{BoostLU} \circ \Phi)(\vx) \,\, .
\end{equation}
From now on, we will consider three different scenarios for applying \boostrelu in multi-label classification.


\subsection{Usage 1: BoostLU in inference only}
\label{sec:usage1}

Since the idea of \boostrelu comes from analyzing the CAM of a model which finished training with AN loss, 
we first propose to apply \boostrelu only during the inference phase of that model.
Initially, this model produces low logits for categories whose label is positive. However, applying \boostrelu increases the corrupted attribution scores and produces higher logits. At the same time, boosting effect is not much for categories whose label is negative; therefore, its logits remain almost the same.
As a result, prediction scores are better separated between samples with positive and negative labels, improving average precision.


\subsection{Usage 2: BoostLU in both training and inference}
\label{sec:usage2}
Next, we consider applying \boostrelu during the training phase with AN loss and the inference phase. 
The gradient of logit $g$ with respect to the attribution score on CAM at location $(i, j)$ (i.e., $\mM_{ij}$) then becomes

\begin{equation}
    \frac{\partial g}{\partial \mM_{ij}} = 
    \begin{cases}
        \alpha/HW, & \mM_{ij} \geq 0\\
        1/HW, & \mM_{ij} < 0 \,\, .
    \end{cases}
    \label{eq:dgdm}
\end{equation}
Compared to the case that does not use \boostrelu, where every spatial location gets a uniform gradient of $1/HW$, the locations with positive attribution scores receive gradients boosted by $\alpha$ times. Thanks to the boosted gradients, these locations are encouraged to produce higher attribution scores during training
when the model receives a positive label.
Also, when a true negative label comes in, these locations are encouraged to produce lower attribution scores.

However, in practice, we observe only marginal improvement in model performance.
It is because the boosted gradients have an adverse effect when false negatives come in as input. 
That is, \boostrelu also boosts the wrong direction of gradients from false negatives, which can be easily seen by combining Equation (\ref{eq:dldg}) and (\ref{eq:dgdm}):

\begin{equation}
    \frac{\partial{\mathcal{L}_{-}}}{\partial{\mM_{ij}}} = 
    \frac{\partial{\mathcal{L}_{-}}}{\partial{g}} \cdot \frac{\partial{g}}{\partial{\mM_{ij}}} \,\, 
    \label{eq:fn}
\end{equation}
Note that $\partial{\mathcal{L}_{-}}/\partial{g}$ has a wrong sign for false negatives, and it decreases CAM values.

\subsection{Usage 3: Combination with Large Loss Modification}
\label{sec:usage3}

To alleviate the problem mentioned above, we propose combining our \boostrelu with recent studies \cite{kim2022large, p-asl} that detect and treat suspicious false negatives while training multi-label classification models. We especially adopt three methods, i.e., LL-R, LL-Ct, and LL-Cp \cite{kim2022large}, since they work on several partial label settings.
When these methods are combined with \boostrelu, they suppress the side effects caused by false negatives. As a result, the model can take full advantage of the boosted gradients from the positive labels during training. Moreover, because these combined methods consider samples with relatively high prediction scores among unobserved labels as false negatives, \boostrelu helps the model detect more false negatives by boosting their logit values.




%% file: 05experiments.tex
\section{Experiments}
\label{sec:experiments}

To validate the efficacy of our proposed method, we report our experimental results on two partial label settings: single positive label (\S\ref{sec:single_positive_label}) and large-scale partial label (\S\ref{sec:openimages}). In both sections, we adopt mean Average Precision (mAP) as an evaluation metric and report the performance on a test set using the model weight with the highest mAP in the validation set. We fix our hyperparameters as $\alpha = 5, \,\beta = 0$. Next, we present analysis results on \S\ref{sec:analysis}.

\subsection{Single positive label}
\label{sec:single_positive_label}
\myparagraph{Datasets.}
We target four multi-label classification datasets: PASCAL VOC 2012 \cite{pascalvoc}, MS COCO 2014 \cite{coco}, NUSWIDE \cite{nuswide}, and CUB \cite{cub}. Each dataset is annotated for 20 classes, 80 classes, 81 concepts, and 312 attributes. Since they are fully annotated, we only keep one positive label and drop the rest of the labels for every training image to build a single positive label setting identical to \cite{single_positive_label}.

\myparagraph{Hyperparameter settings.}
For a fair comparison, we set the same search space as \cite{single_positive_label, zhou2022acknowledging}: $\{8, 16\}$ for batch size and $\{10^{-2}, 10^{-3}, 10^{-4}, 10^{-5}\}$ for learning rate.
We train the model for 10 epochs with Adam optimizer \cite{kingma2014adam}.
LL-R, LL-Ct, and LL-Cp \cite{kim2022large} have a hyperparameter $\Delta_{rel}$ that controls the slope of increase in the modification rate.
We set $\Delta_{rel}=0.5$ for LL-R, $0.2$ for LL-Ct, and $0.1$ for LL-Cp, respectively. We set a 10x learning rate for the last 1x1Conv layer.

\myparagraph{Implementation details.}
We also follow identical configurations as \cite{single_positive_label, kim2022large, zhou2022acknowledging}. Specifically, 20\% of the original training set is used for validation. ResNet-50 \cite{resnet} CNN backbone pre-trained on ImageNet \cite{imagenet1} is used as a feature extractor. 
Each image is resized to 448x448 before being fed to CNN, and only random horizontal flipping is used for data augmentation during training.
Note that some categories do not have positive labels in the CUB dataset on a generated single positive label setting.
In these categories, we do not apply \boostrelu when training as the benefit from the boosted gradient becomes weakened.



{
\setlength{\tabcolsep}{5pt}
\renewcommand{\arraystretch}{1.2}
\begin{table}[t]
\centering
\small
\begin{tabular}{ccc||cc}\hline
\boostrelu & \boostrelu & LL-R & \multicolumn{2}{c}{Performance} \\ \cline{4-5}
in inference & in training & in training & VOC & COCO \\ \hline\hline
 &   &   & 86.10  & 64.58\\ \hline
\checkmark  &   &   &  87.31  & 66.27   \\ \hline
\checkmark & \checkmark  &    & 86.73  & 65.33\\ \hline
 &   & \checkmark  & 88.24  & 70.60 \\ \hline
 & \checkmark  & \checkmark  & 87.18  & 68.45 \\ \hline
\checkmark &   & \checkmark  & 88.90  & 70.87 \\ \hline
\checkmark & \checkmark  &\checkmark   & \textbf{89.27}  & \textbf{72.82} \\\hline

\end{tabular}
\caption{\textbf{Ablation study on BoostLU and LL-R.} We test seven combinations of using \boostrelu and LL-R \cite{kim2022large} on VOC and COCO datasets. Training a model with both LL-R and 
\boostrelu and applying \boostrelu during inference shows the best mAP. }
\vspace{-5pt}
\label{tbl:ablation}
\end{table}
}

{
\setlength{\tabcolsep}{5pt}
\renewcommand{\arraystretch}{1.2}
\begin{table}[t]
\centering
\begin{tabular}{|l||cccc|}
\hline
Methods     & VOC & COCO & NUS & CUB \\ \hline\hline
Full Label & 89.42 & 76.78 & 52.08 & 30.90  \\ \hline
AN & 85.89  & 64.92  & 42.27  & 18.31   \\
LS \cite{labelsmoothing-noise} & 87.90  & 67.15  & 43.77  & 16.26         \\
ASL \cite{asl}    & 87.76  & 68.78  & 46.93  & 18.81       \\ 
ROLE \cite{single_positive_label}  & 87.77  & 67.04  & 41.63  & 13.66  \\
ROLE + LI \cite{single_positive_label}  & 88.26 & 69.12 & 45.98 & 14.86 \\
EM \cite{zhou2022acknowledging}   & 89.09  & 70.70  & 47.15    & 20.85 \\
EM + APL \cite{zhou2022acknowledging}   & 89.19  & 70.87  & 47.59    & \textbf{21.84} \\\hline
LL-R \cite{kim2022large}     & 88.27        &    70.70     & 48.76        &    19.56    \\
\,\, + BoostLU (Ours)& \textbf{89.29}        &    \textbf{72.89}     & \textbf{49.59}        &    19.80    \\
LL-Ct \cite{kim2022large}  & 87.79  & 70.29  & 48.08  & 19.06  \\
\,\, + BoostLU (Ours)& 88.61  & 71.78  & 48.37  & 19.25  \\
LL-Cp \cite{kim2022large}   & 87.44 & 70.27   & 47.92   & 19.21   \\
\,\, + BoostLU (Ours)& 87.81 & 71.41   & 48.61   & 19.34   \\\hline

\end{tabular}
\caption{\textbf{Experimental results on various datasets with single positive label setting.} 
Each number shows the average of mAP in three experiments.
A bold number means the best performance. Results of methods except for LL-R, LL-Ct, and LL-Cp are taken from \cite{zhou2022acknowledging}. We report the reimplemented results for LL-R, LL-Ct, and LL-Cp with the same hyperparameter search space as \cite{zhou2022acknowledging, single_positive_label}.
}
\vspace{-5pt}
\label{tbl:single_positive_label}
\end{table}
}

{
\setlength{\tabcolsep}{5pt}
\renewcommand{\arraystretch}{1.2}
\begin{table*}[t]
\centering
\begin{tabular}{|l||ccccc|c|}
\hline
Methods     & Group 1 & Group 2 & Group 3 & Group 4 & Group 5 & All Classes \\ \hline\hline
CNN-RNN \cite{cnnrnn}   & 68.76  & 69.70  & 74.18  & 78.52 & 84.61 & 75.16        \\ 
Curriculum Labeling \cite{partial_label_2019} & 70.37  & 71.32  & 76.23  & 80.54 & 86.81 & 77.05   \\
IMCL \cite{imcl}  & 70.95 & 72.59 & 77.64 & 81.83 & 87.34 & 78.07  \\
P-ASL \cite{p-asl}   & 73.19  & 78.61  &  85.11    &  87.70 &  90.61 & 83.03  \\
\hline
LL-R \cite{kim2022large}     & 77.76  &   79.07     & 81.94        & 84.51  & 89.36 & 82.53       \\
\,\,\,\,\,\,\,\,  + BoostLU (Ours)    & 79.28  &   80.81     & 83.32        & 85.63  & 90.27 & 83.86       \\
LL-Ct \cite{kim2022large}  & 77.76  & 79.18  & 81.97  & 84.46 & 89.51 & 82.58   \\
\,\,\,\,\,\,\,\, + BoostLU (Ours) & 79.43  & 80.75  &  83.41  & 85.70 & 90.41 & 83.94   \\
LL-Cp \cite{kim2022large}   & 77.49 & 79.22   & 81.89   & 84.51 & 89.18 & 82.46    \\
\,\,\,\,\,\,\,\, + BoostLU (Ours)   &  79.53 &  81.04   &  83.40   &  85.85 &  90.39 &  \bf 84.04    \\
\hline

\end{tabular}
\caption{\textbf{Experimental results on a OpenImages V3 dataset.} Each group includes 1,000 classes without overlapping. Group 1 has the smallest annotations, and Group 5 has the most. The number of annotations increases as the group number increases. 
LL-R, LL-Ct, and LL-Cp are reimplemented and the other results are borrowed from \cite{p-asl}. A bold number shows the best performance.
}
\vspace{-5pt}
\label{tbl:openimages}
\end{table*}
}

\myparagraph{Results of ablation study.}
We first conduct ablation studies on PASCAL VOC and COCO datasets. Its results are reported in Table \ref{tbl:ablation}.
First, we show the performance of the model trained with AN loss in the first row.
In the second row, it can be seen that when \boostrelu is applied during inference of this model, its test performance is improved even without additional training. 
It confirms the property of \boostrelu that compensates for the damaged attribution score.
However, if we further apply \boostrelu while training (third row), the performance improvement is lower than when \boostrelu is applied only during inference. We can observe the side effect of BoostLU that the gradient received by the region with a positive attribution score is boosted even for false negative labels.

In the fourth row, we show the performance of LL-R, which rejects large losses during training. 
We then train the model by applying both LL-R and \boostrelu during training and \boostrelu during inference. Its performance is reported in the final row, and its improvement is much more significant than the case where LL-R is not applied (+0.63 v.s. +1.03 on PASCAL, and +0.75 v.s. +2.22 on COCO). Thanks to LL-R filtering out false negatives, the side effect of the boosted gradient becomes minimized. Moreover, since our \boostrelu helps LL-R detect false negatives, its advantage is further amplified.
We also find in the last three rows that
when we combine \boostrelu with LL-R, applying \boostrelu either during training or during inference results in a performance drop compared to applying it during both phases. This shows that \boostrelu plays a vital role both in training and inference, together with large loss modification methods. 
In particular, it is crucial to apply BoostLU during inference to achieve high performance.
Additional discussion about this is described in the Appendix.

From now on, we will only report the experimental results using the configuration of the last row (\boostrelu in inference + \boostrelu in training + LL-R in training).

\myparagraph{Comparison with prior arts.}
We compare our results with recent state-of-the-art: Label Smoothing (LS) \cite{labelsmoothing-noise}, Asymmetric loss (ASL) \cite{asl}, ROLE (with LinearInit) \cite{single_positive_label}, and Entropy-maximization loss (EM) with Asymmetric Pseudo-Labeling (APL) \cite{zhou2022acknowledging}. We train the network three times and report the average test performance.

The results are shown in Table \ref{tbl:single_positive_label}.
We find that applying \boostrelu in both training and inference consistently improves the performance of LL-R, LL-Ct, and LL-Cp in all datasets, only with little additional computational cost. 
It achieves +1.02, +0.82, and +0.37 mAP improvement in VOC, as well as +2.19, +1.49, and +1.14 mAP improvement in COCO, respectively.
Especially the performance of LL-R + \boostrelu shows the most significant increase, achieving state-of-the-art performance and reaching closest to the full label performance on VOC, COCO, and NUSWIDE.
It also surpasses the previous state-of-the-art method EM+APL which does not use AN assumption on these datasets. 
However, the performance improvement is not that large in CUB.
Since CUB has an annotation for attributes, the number of false negative labels is much higher, and this may increase the side effect of \boostrelu when applied during training. 




\subsection{Large-scale partial label}
\label{sec:openimages}
\myparagraph{Dataset.} We target a partially annotated OpenImages V3 \cite{openimages} dataset which consists of 3.4M training images, 41,620 validation images, and 125,436 test images with 5,000 trainable classes (having more than 30 human-verified samples in the training set and 5 in the valid or test sets). We sort these classes in ascending order by the number of annotations in the training set and divide them into five groups of equal size 1,000. We report the mAP score averaged within each group and the entire 5,000 classes.

\myparagraph{Implementation details.} 
We use ImageNet \cite{imagenet1} pre-trained ResNet-101 \cite{resnet} as a feature extractor, the same as prior works. 
We follow \cite{kim2022large} to set the learning rate as $2 \times 10^{-5}$ and batch size as 288. We train the model for 20 epochs and set $\Delta_{rel} = 0.005$. We resize every image to 224x224 resolution and perform a random horizontal flip during training.
We set a 10x learning rate for the last 1x1Conv layer.

\myparagraph{Results.}
We compare our results with prior works: CNN-RNN \cite{cnnrnn}, Curriculum Labeling \cite{partial_label_2019}, IMCL \cite{imcl}, and P-ASL \cite{p-asl}. 
As shown in Table \ref{tbl:openimages},
\boostrelu also works well in a real partial label scenario.
Combined with LL-R, LL-Ct, and LL-Cp, it boosts their performance by a large margin: improvement of +1.33, +1.36, and +1.58 mAP, respectively. All of the combined methods surpass other previous methods and achieve state-of-the-art performance. 
In particular, LL-Cp + \boostrelu shows the highest 84.04 mAP.


\begin{figure}[t]
    \centering
    \includegraphics[width=\linewidth]{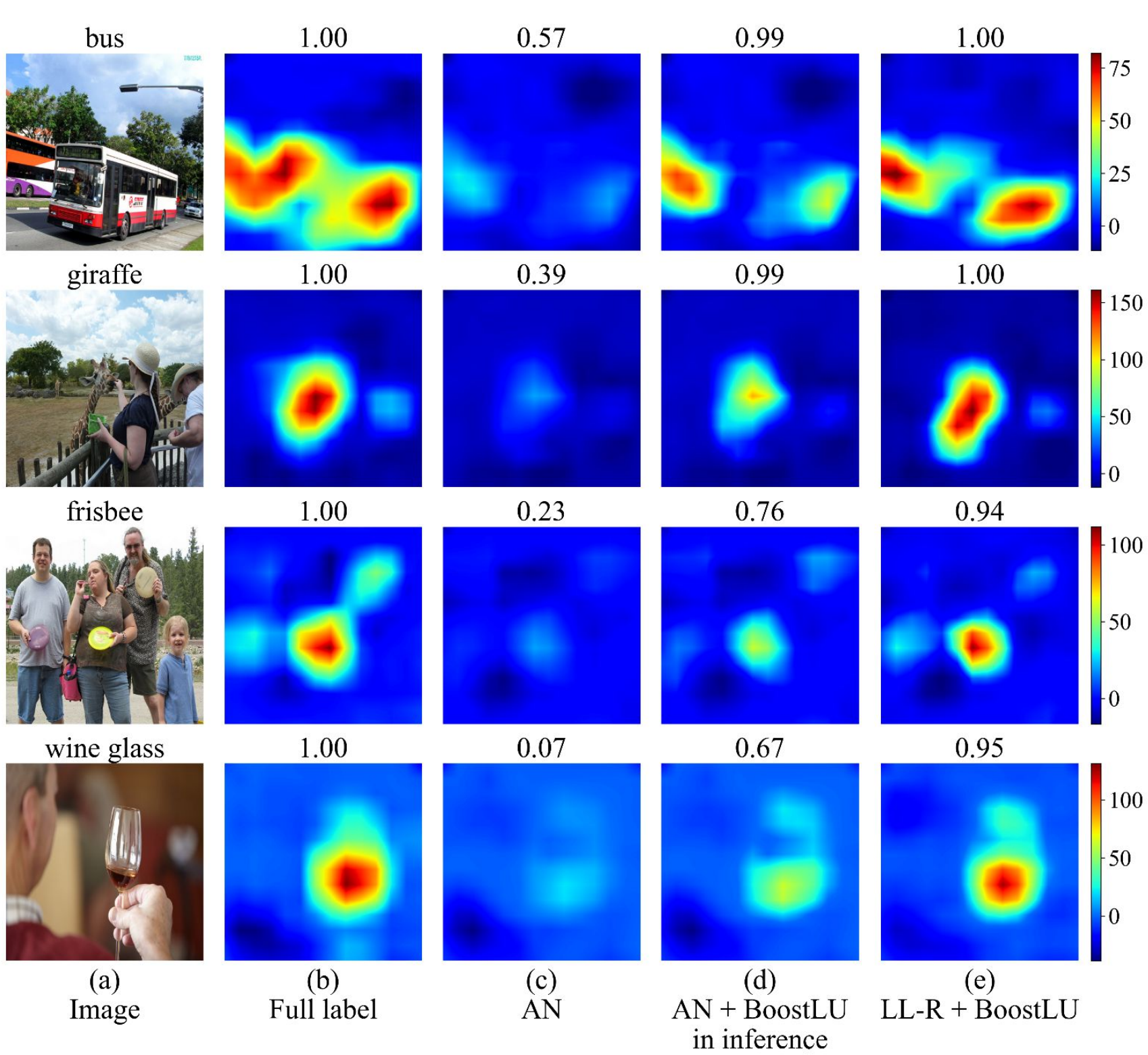}
    \vspace{-20pt}
    \caption{\textbf{Qualitative results.} Categories and their corresponding prediction scores are displayed above the images and CAM results. LL-R + \boostrelu is the closest to the explanation and prediction score of the model trained with full labels.} 
    \vspace{-5pt}
    \label{fig:qualitative}
\end{figure}

\subsection{Analysis}
\label{sec:analysis}

\myparagraph{Qualitative results.} Figure \ref{fig:qualitative} visualizes the CAM results from four different methods.
The category corresponding to the CAM is shown above the image. The prediction score, obtained by averaging attribution scores on CAM and applying sigmoid activation, is shown above each CAM.
First, column (c) shows that a model trained with AN loss gives low prediction scores due to the damage of false negatives.
Although this model highlights similar regions for a given input image, the attribution scores of the corresponding regions are considerably shrunk compared to column (b).

When we perform inference by attaching \boostrelu to this model, it can be seen in column (d) that BoostLU successfully recovers the model's explanation, yielding high prediction scores. For LL-R + \boostrelu in column (e), 
its model explanation is further improved 
due to the role of LL-R and \boostrelu during training which further accelerates the improvement of the attribution score of the highlighted region.
It is the most similar to the explanation of the model trained with full annotation (column (b)) compared to other methods.

\begin{figure}[t]
    \centering
    \includegraphics[width=\linewidth]{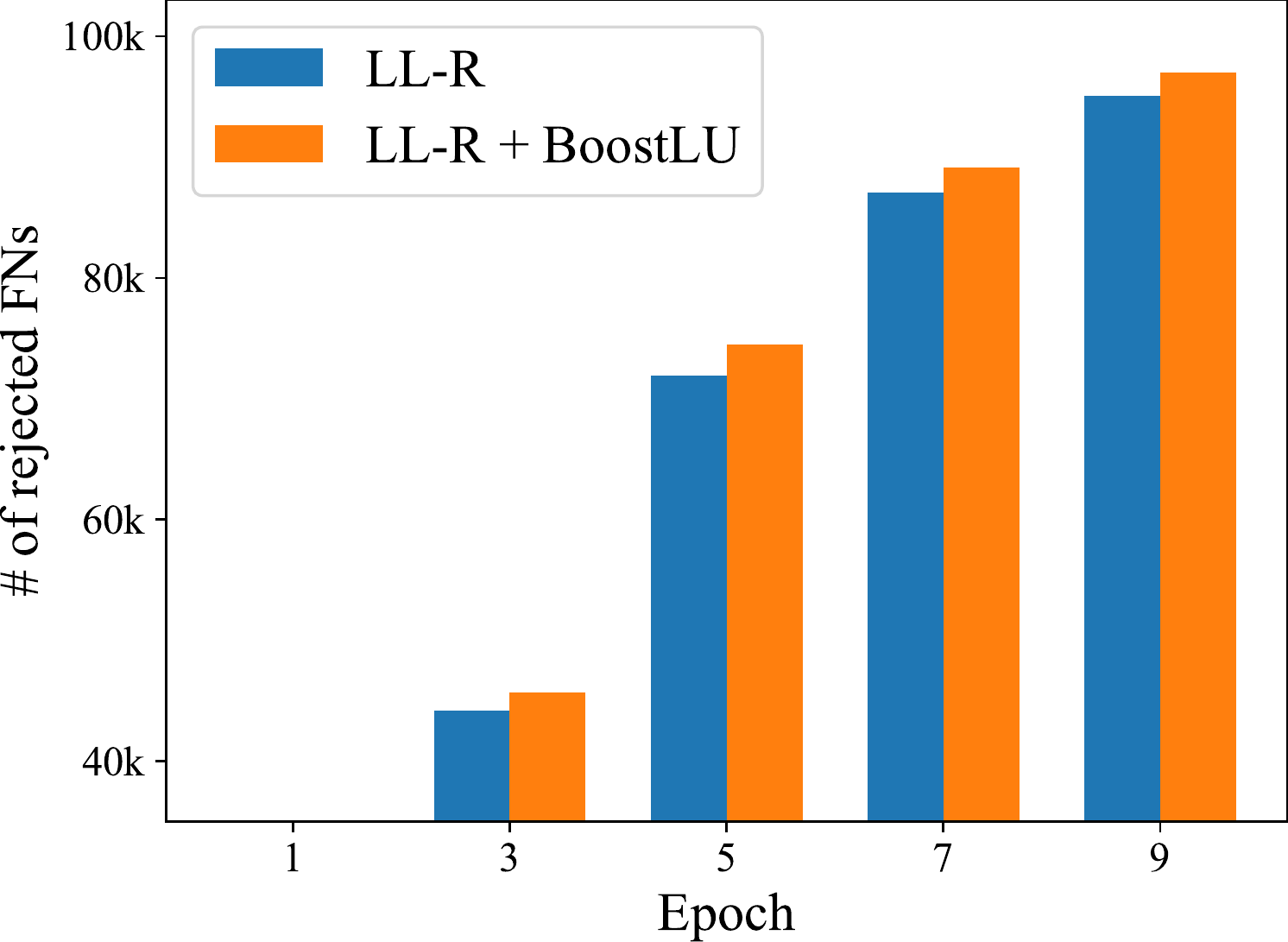}
    \vspace{-15pt}
    \caption{\textbf{Comparison of the number of rejected false negative labels.} \boostrelu helps LL-R detect more false negative labels in every epoch.} 
    \vspace{-5pt}
    \label{fig:numfn}
\end{figure}

\myparagraph{Synergy effect of \boostrelu and large loss modification methods during training.}
We train LL-R and LL-R + \boostrelu on the COCO dataset with the same $\Delta_{rel}=0.5$ to make both models reject the same number of samples during training.
We then inspect how many of the rejected samples are false negative labels.
Figure \ref{fig:numfn} shows the number of false negative labels rejected by each model per epoch.
It can be seen that after the warmup phase (first epoch),
LL-R + \boostrelu rejects more false negatives than LL-R in every epoch. It is because 
\boostrelu boosts the logit value of false negative samples, thus boosting the 
large loss modification methods' ability to detect false negatives.
At the same time, it also reduces the number of true negative samples that the model incorrectly rejects, further contributing to performance improvement.

%% file: 06conclusion.tex
\section{Conclusion}

In this paper, we studied the effect of false negative labels on model explanation when assuming unobserved labels as negatives in a partially annotated multi-label classification situation. 
We found that the overall spatial shape of the explanation tends to be preserved, but the scale of attribution scores is significantly affected.
Based on these findings, we proposed a conceptually simple piece-wise linear function \boostrelu that compensates for the damaged attribution scores.
Through several experiments, we confirmed that \boostrelu successfully contributed to bridging the explanation of the model closer to the explanation of the model trained with full labels.
Furthermore, combined with large loss modification methods, it achieved state-of-the-art performance on several multi-label datasets. \newline

\myparagraph{Acknowledgements.}
Youngwook Kim thanks Junghyun Lee and Youngmin Ro for their valuable help.
Jae Myung Kim thanks the European Laboratory for Learning and Intelligent Systems (ELLIS) PhD program and the International Max Planck Research School for Intelligent Systems (IMPRS-IS) for support.
This work is in part supported
by National Research Foundation of Korea (NRF, 2021R1A4A1030898(10\%)), Institute of Information \& communications Technology Planning \& Evaluation (IITP, 2021-0-00106 (50\%), 2021-0-01059 (20\%), 2021-0-00180 (20\%)) grant funded by the Ministry of Science and ICT (MSIT), INMAC, and BK21-plus. 
This work is also supported by DFG project number 276693517, by BMBF FKZ: 01IS18039A, by the ERC (853489 - DEXIM), and by EXC number 2064/1 – project number 390727645.